\documentclass[sigconf]{acmart}

\usepackage{booktabs} 
\usepackage{colortbl}
\usepackage{amssymb}
\usepackage{color}
\usepackage{graphicx}

\usepackage{pifont}

\renewcommand\footnotetextcopyrightpermission[1]{} 
\newcommand\given[1][]{\:#1\vert\:}

\usepackage{mathtools}

\begin{document}
\title{Evaluating Fairness Metrics in the Presence of Dataset Bias}

\author{J. Henry Hinnefeld}
\affiliation{%
  \institution{Civis Analytics}
}
\email{hhinnefeld@civisanalytics.com}

\author{Peter Cooman}
\affiliation{%
  \institution{Civis Analytics}
}
\email{pcooman@civisanalytics.com}

\author{Nat Mammo}
\affiliation{%
  \institution{Civis Analytics}
}
\email{nmammo@civisanalytics.com}

\author{Rupert Deese}
\affiliation{%
  \institution{Civis Analytics}
}
\email{rdeese@civisanalytics.com}

\begin{abstract}
Data-driven algorithms play a large role in decision making across a variety of industries. Increasingly, these algorithms are being used to make decisions that have significant ramifications for people's social and economic well-being, e.g. in sentencing, loan approval, and policing. Amid the proliferation of such systems there is a growing concern about their potential discriminatory impact. In particular, machine learning systems which are trained on biased data have the potential to learn and perpetuate those biases. A central challenge for practitioners is thus to determine whether their models display discriminatory bias. Here we present a case study in which we frame the issue of bias detection as a causal inference problem with observational data. We enumerate two main causes of bias, sampling bias and label bias, and we investigate the abilities of six different fairness metrics to detect each bias type. Based on these investigations, we propose a set of best practice guidelines to select the fairness metric that is most likely to detect bias if it is present. Additionally, we aim to identify the conditions in which certain fairness metrics may fail to detect bias and instead give practitioners a false belief that their biased model is making fair decisions.
\end{abstract}

\maketitle

\section{Introduction}

Data-driven algorithms are used to inform decision making in a growing number of socially impactful domains, including criminal justice, finance, education, and hiring \cite{oneil2016weapons}. The increasing influence of these systems has prompted a corresponding increase in public concern over their potential to produce discriminatory outputs. There is ample evidence that such concerns are well founded: algorithmic models used to generate criminal recidivism predictions \cite{larson2016we, chouldechova2017fair}, employment advertising \cite{datta2015automated}, and web product pricing \cite{staples2012} have all been shown to discriminate based on protected attributes (e.g. race or gender).  

Discriminatory behavior is not a deliberate feature of these systems, but rather the result of biases present in the input data used to train the systems \cite{calders2013unbiased}. A key hurdle for industrial applications of machine learning models is thus to determine whether the raw input data used to train the model contains discriminatory bias. This question is not straightforward: there are many ways to quantify bias, and many subtleties to consider when interpreting the results of such measurements. In light of this difficulty, we present a case study which examines the ability of six different fairness metrics to detect unfair bias in predictions generated by models trained on datasets containing known, artificial bias.

Our contributions are threefold. First, we frame the problem of bias detection as a causal inference problem with observational data. This framing highlights the subtleties that accompany causal studies using observational data, and emphasizes the parallels between those challenges and the difficulties associated with measuring fairness in machine learning settings. Second, we investigate the performance of six different fairness metrics under conditions of varying dataset bias. Specifically, we examine the consequences of making conclusions based on these metrics in the presence of uncertainty about the causal origins of bias in the dataset. Finally, based on these potential consequences we present a set of recommended best practices to guide fairness metric selection.

\section{Related Work}

The literature on fairness in machine learning is broadly categorized by three (often overlapping) goals: to quantify the degree of unfair bias present in data or model predictions \cite{feldman2015certifying,hardt2016equality,zliobaite2015survey,dwork2012fairness,romei2014multidisciplinary}, to remove unfair bias from data or model predictions \cite{feldman2015certifying, zemel2013learning}, and to develop machine learning algorithms which include fairness constraints \cite{zemel2013learning, kamishima2012fairness, zafar2015learning, calders2010three}. A primary question for all three approaches is: ``How should `fairness' be defined?'' 

This question is the subject of active debate. Several metrics have been suggested which attempt to mathematically define various competing notions of fairness \cite{dwork2012fairness, feldman2015certifying, zliobaite2015survey, zliobaite2015relation, hardt2016equality, chouldechova2017fair}; the proliferation of these metrics reflects the many and sometimes mutually exclusive \cite{kleinberg2016inherent} interpretations of `fairness' in machine learning contexts. Here we focus on six metrics which appear repeatedly in the literature: Difference in Means \cite{zliobaite2015survey}, Difference in Residuals \cite{zliobaite2015survey}, Equal Opportunity \cite{hardt2016equality}, Equal Mis-opportunity \cite{hardt2016equality}, Disparate Impact \cite{feldman2015certifying}, and Normalized Mutual Information \cite{zliobaite2015survey}.

Certain definitions of fairness, and thus certain fairness metrics, are demonstrably inappropriate in particular circumstances. For example, if there is a legitimate reason for a difference in the rate of positive labels between members of different protected classes (e.g. incidence of breast cancer by gender) then statistical parity between model results for the protected classes would be an inappropriate measure. More generally, metrics are inappropriate when they enforce an equality which is inconsistent with ground truth. However, in practical, real-world machine learning settings the only available data may contain an unfairly biased representation of ground truth. This situation presents a conundrum: in order to select an appropriate measure of bias for a given dataset one must first know the bias in that dataset. Here we address this conundrum by considering the consequences of selecting a fairness metric based on a mistaken assessment of the types of bias present in a given dataset.

\section{Bias Detection through the lens of Observational Causal Inference}

We frame the problem of detecting unfair bias in a machine learning setting as a causal inference problem with observational data. This framing serves two purposes: first, it allows us to enumerate distinct causal origins of dataset bias and to evaluate the performance of different fairness metrics in different regions of `dataset bias space'.  Second, it highlights the relationship between the shortcomings of existing fairness metrics and the difficulties associated with causal inference on observational data. To be clear, in this work we are not applying causal inference techniques to detect bias. Rather, we are evaluating existing fairness metrics from the literature within a framework inspired by observational causal inference.

In this study we distinguish between two types of dataset bias, which we call `sample bias' and `label bias'. In this terminology, `sample bias' refers to the case when the sampling process which generates the data is not uniform across protected classes and outcome labels. For example, consider a dataset consisting of applicants to a graduate school. Certain academic disciplines have a large gender disparity in applicants. If department selectivity is correlated with applicant gender disparity then a dataset of applicants to such a program would contain sample bias because the sampling process would preferentially generate e.g., men who are more likely to be accepted and women who are more likely to be rejected. 

We define `label bias' as the case when there is a causal link between a protected attribute and the class label assigned to an individual which is not warranted by ground truth. Consider a dataset composed of elementary school students, with a dependent variable that indicates whether the student misbehaves. Studies have shown that Black and Latinx children are more likely than White children to receive suspensions or expulsions for similar problem behavior \cite{skiba2011we}, so if our dataset's dependent variable were based on suspensions it would contain label bias. This taxonomy of bias types is consistent with other classifications presented in the literature, e.g.  \cite{calders2013unbiased}, however we distinguish the definitions presented here by their emphasis on the causal origin of the bias.  

Several pitfalls of measuring fairness can be understood through the lens of causal inference. First, consider situations which display Simpson's paradox \cite{blyth1972simpson}, i.e. cases where different levels of data aggregation produce different fairness conclusions. An analysis of graduate admissions data from Berkeley offers one such case \cite{bickel1975sex}, in which the aggregate data show an admissions bias against women, but when the data are disaggregated to the department level the bias is reversed. This difficulty stems from the causal influence of the protected class (in this case gender) on the presence of an individual in the set of applicants to each department: women preferentially apply to departments with lower acceptance rates. The fairness question ``Does a person's gender cause him/her to be more likely to be accepted?'' is confounded by the causal influence of the person's gender on the sampling process which generated the dataset. Simpson's paradox may manifest in fairness measurements whenever there is a causal link between a protected attribute and the sampling process which generates the dataset, i.e. whenever there is sample bias.

Next, consider the case where a dataset contains label bias, i.e. when there is an unwarranted causal relationship between a protected class and the label assigned to members of that class. Models trained on such a dataset may produce unfairly biased predictions even when the sensitive attribute is omitted due to collinearities between the protected class and other explanatory variables \cite{calders2013unbiased}. Historic ``redlining'' \cite{zliobaite2011handling, kamishima2012fairness} practices, in which zip codes were used as a proxy for race in mortgage lending decisions are an example of a deliberate application of this effect, however the same phenomenon occurs even in the absence of malicious intent if machine learning models are applied naively to datasets containing label bias. Detecting label bias requires determining the effect of a protected attribute on a model's predictions in the presence of other correlated variables. This is a classic causal inference problem, and is plagued by the same complications that accompany causal studies using observational data, e.g. omitted, included, or incomplete variable bias \cite{romei2014multidisciplinary}.

\section{Case Study}

Selecting an appropriate fairness metric for a given dataset is a chicken-and-egg problem: the types of bias present in the dataset determine which metric is appropriate, but determining which types of bias are present requires some way to measure bias. Here we approach this problem by evaluating the performance of six different metrics on datasets containing known, artificial bias.

\subsection{Experimental Methods}

To investigate the performance of the fairness metrics we perform two experiments. We begin with a dataset containing demographic information about a subset of U.S. citizens, with a dependent variable that indicates the likelihood (on a scale from 0 to 1) that each person will sign up for the services of an unspecified state agency\footnote{Contractual limitations prevent us from identifying the client whose data we used for this analysis or making the data available.}. This original dataset is summarized in Table \ref{tbl:original}.

For Experiment A, we create a new base dataset where ground truth positive rates and class membership are both balanced. This is done by selecting only the white citizens in the original dataset and then randomly re-assigning race labels. For Experiment B, we use the unmodified original dataset as the new base dataset. In both experiments we define the base dataset to be ground truth; note that in Experiment B this means that the ground truth positive rate differs between groups.

Each experiment proceeds by introducing artificial causal bias to the relevant base dataset, splitting the resulting biased data into training and testing subsets, training an elastic net logistic classification model on the training set, scoring the model on the test set, and then applying each fairness metric to the model outputs. This process is repeated for each possible combination of bias types, as summarized in Table \ref{tbl:datasets}. Finally, we evaluate each metric on its ability to detect the artificially introduced bias.  

\begin{table}[t]
  \centering
  \begin{tabular}{ l | r | r | r }
      			   & $Y = 1$      & $Y = 0$         & \\ \hline
    race = black & 1,296 & 9,357 & 10,653 \\ \hline
    race = white & 64,536 & 54,804 & 119,340 \\ \hline
    & 65,832 & 64,161 & 129,993
    
  \end{tabular}
  \caption{\textmd{The original dataset contains both imbalanced classes and differing rates of positive labels between protected class. Note that the original dependent variable is a likelihood score between 0 and 1; here we present summary statistics by assigning binary labels based on a score threshold of 0.5.
}}
  \label{tbl:original}
\end{table}

We note one important detail of this method: in this analysis we are treating the likelihood score from our original dataset as ground truth. This score is itself modeled (prior to our experiments here), and thus subject to causal sample and or label bias. However, our conclusions are generally applicable to situations where the ground truth contains imbalances similar to those in our base datasets, i.e. balanced classes with equal positive rates as in Experiment A, or imbalanced classes with differing positive rates as in Experiment B.

To introduce causal label bias into the training data we assign different label thresholds based on race. Specifically, in our artificially biased datasets we define:
$$ 
\widetilde{Y} =
\left\{
	\begin{array}{ll}
		1 \; \text{if score} \geq 0.3, \; \text{else} \; 0  & \mbox{if race = white} \\
		1 \; \text{if score} \geq 0.7, \; \text{else} \; 0  & \mbox{if race = black}
	\end{array}
\right.
$$
where $\text{score}$ is the likelihood score from the original dataset. We also define an unbiased label:
$$
Y' = 1 \; \text{if score} \geq 0.5, \; \text{else} \; 0
$$
To introduce causal sample bias we preferentially sample white citizens having higher scores, while sampling black citizens uniformly:
\begin{equation*}
\widetilde{P}(x \in X) =
\left\{
	\begin{array}{ll}
		0.8  & \mbox{if race = white and score} \geq 0.5  \\
		0.2  & \mbox{if race = white and score} < 0.5 \\
		1  & \mbox{if race = black} \\
	\end{array}
\right.
\end{equation*}
where $P(x \in X)$ is the probability a given person in the base dataset is included in the artificial training dataset. We also define an unbiased sampling process:
$$
P'(x \in X) = 0.5
$$

For each experiment, we train four elastic net logistic classification models, one on each of the four datasets in Table \ref{tbl:datasets}, and then apply the following metrics to the model predictions, where $\widehat{Y}$ is the predicted label, $Y$ is the training label (either $\widetilde{Y}$ in the case of label bias or $Y'$ in the non-label biased case) and $S = 1$ indicates membership in the protected class (in this case $\text{race = black}$):

\begin{enumerate}

  	\item Difference in mean scores \cite{zliobaite2015survey}:
    $$ 
    E \left\{ \widehat{Y} \given S = 1 \right\} - E \left\{ \widehat{Y} \given S = 0 \right\}
    $$
    
    \item Difference in average model residuals \cite{zliobaite2015survey}:
    $$
    E \left\{ \widehat{Y} - Y \given S = 1 \right\} - E \left\{ \widehat{Y} - Y \given S = 0 \right\}
    $$
    
	\item Equal Opportunity \cite{hardt2016equality}:
    $$
      \text{Pr}\left\{ \widehat{Y} = 1 \given S = 1, Y = 1 \right\} 
    - \text{Pr}\left\{ \widehat{Y} = 1 \given S = 0, Y = 1 \right\}
    $$
    
    \item Equal Mis-Opportunity \cite{hardt2016equality}:
    $$
      \text{Pr}\left\{ \widehat{Y} = 1 \given S = 1, Y = 0 \right\}
    - \text{Pr}\left\{ \widehat{Y} = 1 \given S = 0, Y = 0 \right\}
    $$
    Note that equal opportunity and equal mis-opportunity together comprise the Equal Odds fairness criterion \cite{hardt2016equality}.
    
	\item Disparate Impact \cite{feldman2015certifying}:
    $$
    \frac{\text{Pr} \left\{\widehat{Y} = 1 \given S = 1 \right\}}{\text{Pr} \left\{\widehat{Y} = 1 \given S = 0 \right\}}
    $$
    
	\item Normalized Mutual Information score \cite{zliobaite2015survey}:
    $$
    \begin{aligned}
    \frac{1}{\sqrt{H(\widehat{Y}) H(S)}} 
    \sum\limits_{\widehat{Y}, S} \text{Pr} \left\{\widehat{y}, s \right\} 	 
    \log
    \frac{\text{Pr} \left\{\widehat{y}, s \right\}} 
    	{\text{Pr} \left\{\widehat{y} \right\} \text{Pr} \left\{s \right\}},
    \\
    H(Y) = - \sum\limits_Y \text{Pr} \left\{y\right\} \log \text{Pr} \left\{y \right\} 
    \end{aligned}
	$$
    
\end{enumerate}

\begin{table}[t]
  \centering
  \begin{tabular}{ l | c | c | }
      			   & No Label Bias      & Label Bias         \\ \hline
    No Sample Bias & \textbf{Dataset 1} & \textbf{Dataset 3} \\ \hline
    Sample Bias    & \textbf{Dataset 2} & \textbf{Dataset 4} \\
    \hline
  \end{tabular}
  \caption{\textmd{Datasets prepared with artificially introduced causal bias. Each dataset is generated from a base dataset according to the sampling and labeling procedures described in the text.}}
  \label{tbl:datasets}
\end{table}

\subsection{Results and Discussion}

\begin{figure}[t]
\includegraphics[width=\linewidth]{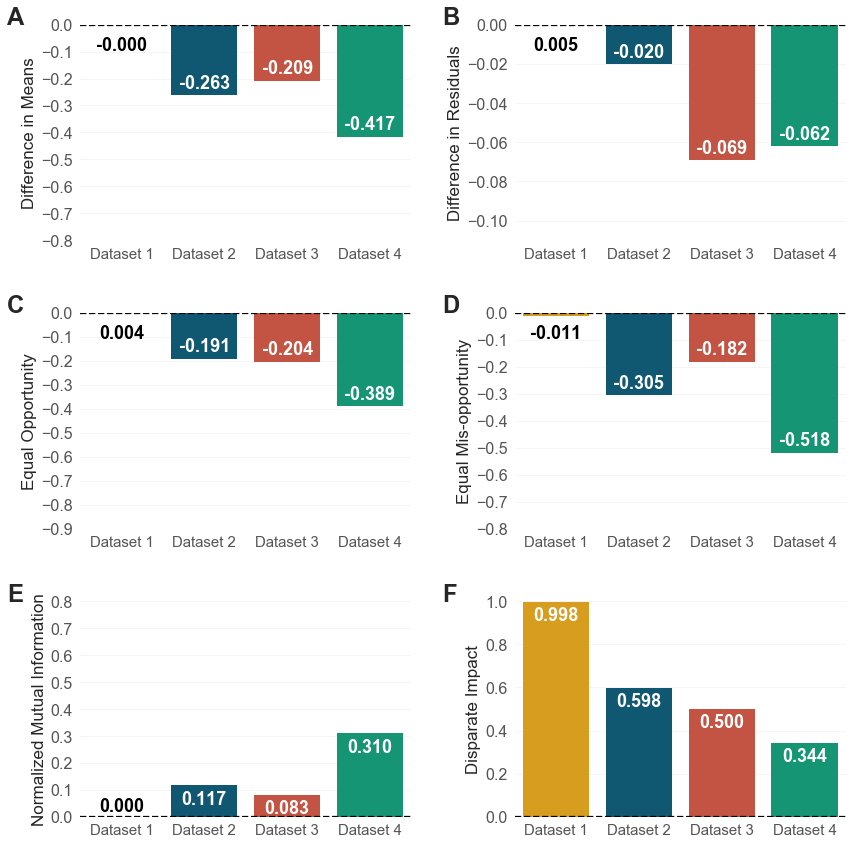}
\caption{\textmd{
Results of Experiment A. Each fairness metric is applied to models trained on datasets containing causal bias. Dataset definitions are as in Table \ref{tbl:datasets} and metric definitions are in the text. For Figures A-E a value of 0 represents non-discrimination, for Figure F a value of 1 represents non-discrimination. 
}}
\label{fig:balanced}
\end{figure}

\begin{figure}[t]
\includegraphics[width=\linewidth]{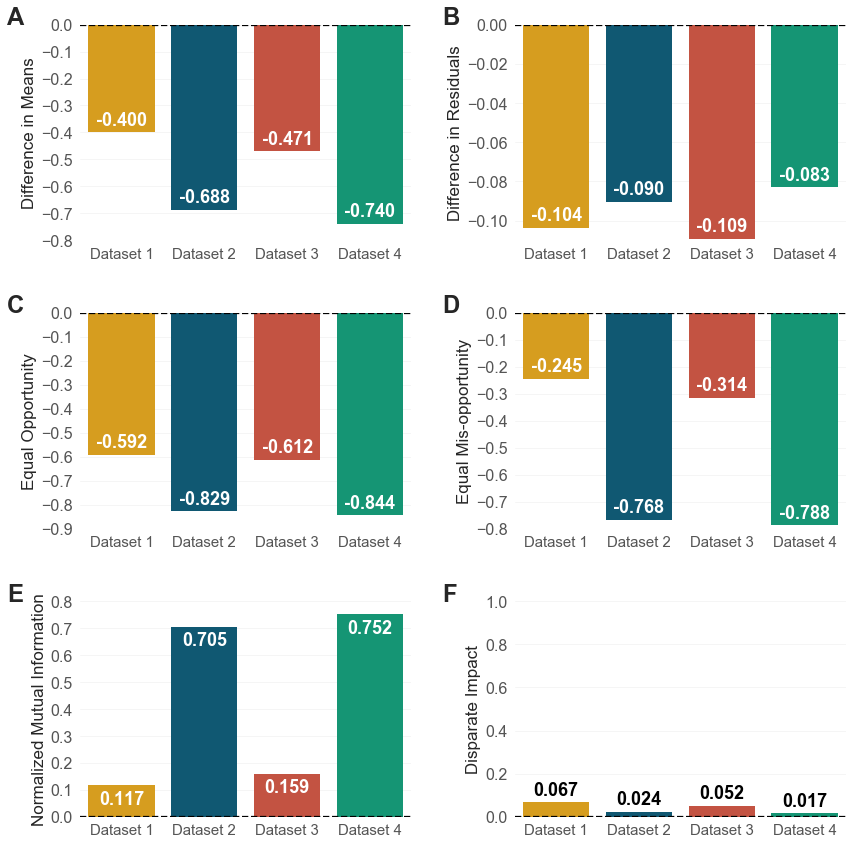}
\caption{\textmd{
Results of Experiment B. All definitions are the same as in Figure \ref{fig:balanced}. 
}}
\label{fig:nonbalanced}
\end{figure}

The results of Experiments A and B are presented in Figures \ref{fig:balanced} and \ref{fig:nonbalanced}, respectively. With the exception of Disparate Impact, there are no established thresholds for determining what level of measured bias constitutes an unfair model. Therefore, we focus on comparisons, both within and between experiments, to illustrate the performance of each metric under different bias and ground truth conditions. 

In Experiment A all metrics correctly identify Dataset 1 as least biased, and all metrics except Difference in Residuals correctly identify Dataset 4 as most biased. Comparing Datasets 2 and 3 shows that most metrics display similar sensitivities to both types of bias, with the exception of Difference in Residuals which is more sensitive to label bias, and Equal Mis-opportunity which is more sensitive to sample bias.

In Experiment B all metrics except Difference in Residuals again correctly identify Dataset 1 as least biased, however all metrics also detect significant bias in Dataset 1 -- recall that Dataset 1 contains no artificially introduced bias and is representative of ground truth. Comparing Datasets 2 and 3 shows that in Experiment B most metrics display greater sensitivity to sample bias than to label bias. In fact, comparing Datasets 1 and 3 shows that most metrics display minimal sensitivity to label bias.  

Several empirical conclusions emerge from the results of these experiments. First, the inability of all metrics tested here to distinguish between bias and legitimate imbalances in the ground truth positive rate illustrates the importance of considering the expected ground truth. Second, the dependence of metric sensitivity on bias type, and the insensitivity of most metrics to label bias in Experiment B, illustrates the importance of considering the causal origin of the bias in a dataset. Finally, in practical applications fairness metrics are applied to a single dataset, and the resulting value must be interpreted alone without comparing against a known unbiased result. However, metric values depend strongly on the imbalances present in ground truth which makes interpreting an individual metric difficult without some external context.

From these observations we conclude that no single fairness metric is universally applicable. When evaluating fairness in machine learning settings practitioners must carefully consider both the imbalances which may be present in the ground truth they hope to model, and the origins of the bias in the datasets they will use to create those models. We end by proposing a set of best practices to guide practitioners when evaluating which fairness metric to use.

\section{Best Practice Guidelines}

Having an a-priori expectation for ground truth positive rates between classes crucially informs which fairness metrics are appropriate, and how their values should be interpreted. In the case where external legal or moral considerations require that the positive rates be equal most metrics can be applied and interpreted in a straightforward manner. Conversely, in the case where ground truth positive rates differ between classes interpreting the results of fairness metrics is difficult.

In the difficult, imbalanced case causal reasoning about the data collection and labelling procedures can inform metric selection. Specifically, if the data collection is susceptible to sample bias then Normalized Mutual Information is a reasonable metric, as it displays good sensitivity to sample bias when ground truth positive rates are imbalanced. Detecting label bias in the imbalanced case is extremely challenging. Additionally, Disparate Impact is particularly ill-suited to the imbalanced case.

Finally, our key recommendation for practitioners is that absent an external source of certainty about ground truth, fairness metrics in machine learning must be interpreted with a healthy dose of human judgement.

\bibliographystyle{ACM-Reference-Format}
\bibliography{main} 

\end{document}